\documentclass[conference,a4paper]{IEEEtran}
\IEEEoverridecommandlockouts

\usepackage{cite}
\usepackage{amsmath,amssymb,amsfonts}
\usepackage{microtype}
\usepackage{url}
\usepackage{xurl}
\usepackage{booktabs,array}
\usepackage{calc}
\usepackage[hidelinks,breaklinks=true]{hyperref}
\hypersetup{
  pdftitle={DACRI: Decision-Aware Causal Intervention Ranking for Critical Supply Chains},
  pdfauthor={Shiqi Huang; Jiani He; Dingyan Shang; Yihua Xu; Jize Li; Yan Lyu; Lakshmi Muraleedharan Nair},
  pdfsubject={Author-accepted manuscript for ICECCME 2026},
  pdfkeywords={supply chain resilience, causal discovery, anomaly detection, intervention ranking, learning-to-rank, decision support, critical infrastructure}
}
\providecommand{\real}[1]{#1}
\providecommand{\tightlist}{\setlength{\itemsep}{0pt}\setlength{\parskip}{0pt}}
\def\BibTeX{{\rm B\kern-.05em{\sc i\kern-.025em b}\kern-.08em
    T\kern-.1667em\lower.7ex\hbox{E}\kern-.125emX}}

\begin{document}

\title{DACRI: Decision-Aware Causal Intervention Ranking for Critical Supply Chains}

\author{%
\IEEEauthorblockN{%
Shiqi Huang\textsuperscript{*},
Jiani He\textsuperscript{*},
Dingyan Shang\textsuperscript{*},
Yihua Xu,
Jize Li,
Yan Lyu,
Lakshmi Muraleedharan Nair}
\IEEEauthorblockA{%
Independent Researcher \\
\{juliahuangsq01, dingyanshang, yxu442, lmuraleedharanair\}@gmail.com,\\
jianihe@alum.mit.edu, lyu.yan@northeastern.edu, jizel@bu.edu}}

\maketitle
\begingroup
\renewcommand{\thefootnote}{*}
\footnotetext{Shiqi Huang, Jiani He, and Dingyan Shang contributed equally to this work.}
\endgroup
\begingroup
\renewcommand{\thefootnote}{}
\footnotetext{\scriptsize
This is the authors' accepted manuscript for presentation at the
International Conference on Electrical, Computer, Communications and
Mechatronics Engineering (ICECCME 2026), 15--17 October 2026, Bali,
Indonesia. \copyright~2026 IEEE. Personal use of this material is permitted.
Permission from IEEE must be obtained for all other uses, in any current or
future media, including reprinting or republishing this material for
advertising or promotional purposes, creating new collective works, for resale
or redistribution to servers or lists, or reuse of any copyrighted component
of this work in other works.}
\endgroup

\begin{abstract}
Detecting or attributing a supply-chain disruption is not the same as
selecting the intervention that maximizes recoverable net value. We
present CriticalSCM-Bench v1, a controlled synthetic benchmark with
causal ground truth, paired factual/counterfactual rollouts, and an
explicit net-value objective. Relative to a full-information
train-selected static benchmark, LambdaMART improves median normalized
net value by 5.7--16.2\%, with paired statistical support on the semiconductor and
critical-material archetypes but not on digital infrastructure. On
digital infrastructure, a domain-informed constant-buffer policy
remains stronger, showing that greater model complexity is not uniformly
justified. Across partial and delayed settings, LambdaMART retains 33--75\%
of full-clamp value. Stress tests further show that intervention fidelity,
timing, cost, and held-out disruptions can alter policy ordering. Critical
materials show the weakest out-of-distribution retention. Separately, a
guarded explanation study over 540 generations preserves every fixed
intervention decision after deterministic validation and template fallback,
although exact wording remains unstable. Within this controlled setting, the
results identify regimes in which adaptive ranking adds value and those in
which simpler structural policies remain preferable.
\end{abstract}


\section{Introduction}

Small upstream signals can cascade into shortages before downstream risk
becomes visible \cite{b1,b2,b3}. Anomaly detection answers ``what is
abnormal?'' Root-cause analysis asks ``where did it begin?'' The
technical gap is not detection alone. It is deciding where an operator
should act now to recover the most value.

DACRI addresses this gap by framing intervention selection as a
learning-to-rank problem. The objective combines counterfactual loss
avoided with the cost of acting. Our contributions follow.

\begin{enumerate}
\def\labelenumi{\arabic{enumi}.}
\tightlist
\item
  \textbf{CriticalSCM-Bench v1}: a controlled benchmark with
  intervention costs and paired counterfactual labels for three
  critical-supply-chain archetypes.
\item
  \textbf{Decision-aware evaluation}: a net-value objective and a baseline
  hierarchy spanning information-light heuristics, attribution-as-action,
  a full-information static benchmark, an oracle-like structural rule,
  learned ranking, and a perfect-value upper bound.
\item
  \textbf{Assumption stress tests}: partial/delayed interventions,
  four feasibility specifications, cost and detection timing, learning
  curves, held-out disruptions, and an oracle ceiling.
\end{enumerate}

A secondary guarded-explanation study evaluates whether a fixed DACRI
decision can be rendered into operator-facing text without allowing the
language model to own or alter the recommendation.

\section{Related Work}

\textbf{Supply-chain and inventory models.} Ripple-effect work traces
disruption propagation \cite{b13}, while multi-tier visibility remains
incomplete \cite{b23,b24}. METRIC \cite{b17} optimizes buffers for known
topology and demand/lead-time distributions; DACRI instead ranks an
episode-specific action after an anomaly. Constant-buffer tests the added
value of learning beyond this downstream-buffer insight. Related data-driven
work we are aware of addresses forecasting and planning under demand
uncertainty---mixed-frequency production planning \cite{b33}, newsvendor
ordering under shifting demand \cite{b34}, and throughput forecasting
\cite{b35}---rather than selecting where to act once a disruption is already
propagating.

\textbf{Anomaly detection.} Isolation Forest \cite{b5}, long short-term
memory (LSTM) autoencoders, USAD \cite{b6}, and MTAD-GAT \cite{b7} capture
multivariate patterns. Microsoft's deployed service reports lead time as
the customer-visible quality measure \cite{b21}.

\textbf{Alert triage and prioritization.} Severity- or anomaly-magnitude
ranking is a common information-light default across operational domains, from
security operations \cite{b25} to multilevel anomaly detection \cite{b26}.
A recurring finding is that severity alone is insufficient and that
prioritization should reflect downstream or business impact rather than
deviation size \cite{b27,b28,b29}. Supply chains lack a corresponding
benchmark with causal ground truth and explicit intervention cost; DACRI
evaluates impact as counterfactual net value.

\textbf{Temporal causal discovery.} Granger causality \cite{b14} is a
linear baseline. PCMCI/PCMCI+ \cite{b15} estimates lagged relationships
under autocorrelation. DYNOTEARS \cite{b16} frames structure learning as
continuous optimization. DACRI treats the graph as an intermediate
consumed by the ranker.

\textbf{Treatment-effect estimation.} Meta-learners estimate heterogeneous
effects from observational data \cite{b36}. DACRI instead uses paired
simulator counterfactuals to rank cost-adjusted feasible actions; it does not
estimate a calibrated treatment-effect surface.

\textbf{Guarded language-model explanations.} Retrieval-grounded generation
provides evidence-conditioned context \cite{b30}, while unsupported claims
remain a concern in hallucination detection and factuality evaluation
\cite{b31,b32}. DACRI fixes the intervention upstream and measures decision
preservation and packet-level grounding, not general faithfulness.

\section{Problem Setting}

\subsection{Archetypes and KPI Schema}

Key performance indicators (KPIs) define the observable dimensions of
each archetype.

Three archetypes support full evaluation: \textbf{digital infrastructure
hardware lifecycle (DI)}, calibrated using drive-reliability and electronics
lifecycle data \cite{b8,b9};
\textbf{semiconductors (SE)} (UCI SECOM \cite{b10}), and \textbf{critical
materials (CM)} (USGS \cite{b11}).

Table~\ref{tab:kpi} summarizes the aligned KPI schema across the three
supply-chain archetypes.

\begin{table*}[!tbp]
\caption{KPI schema for the three critical supply-chain archetypes.}
\label{tab:kpi}
\centering\footnotesize\setlength{\tabcolsep}{3.5pt}
\begin{tabular}{@{}
  >{\raggedright\arraybackslash}p{(\textwidth - 6\tabcolsep) * \real{0.1774}}
  >{\raggedright\arraybackslash}p{(\textwidth - 6\tabcolsep) * \real{0.2419}}
  >{\raggedright\arraybackslash}p{(\textwidth - 6\tabcolsep) * \real{0.2581}}
  >{\raggedright\arraybackslash}p{(\textwidth - 6\tabcolsep) * \real{0.3226}}@{}}
\toprule
\begin{minipage}[b]{\linewidth}\raggedright
Dimension
\end{minipage} & \begin{minipage}[b]{\linewidth}\raggedright
Digital Infra.
\end{minipage} & \begin{minipage}[b]{\linewidth}\raggedright
Semiconductors
\end{minipage} & \begin{minipage}[b]{\linewidth}\raggedright
Critical Materials
\end{minipage} \\
\midrule

Demand & repair demand & production demand & refined-material demand \\
Supply & spare availability & component capacity & material
availability \\
Inventory & spare buffer & component inventory & stockpile buffer \\
Lifecycle & recovery yield & rework yield & recycler yield \\
Quality & fault rate & production yield & purity rate \\
Service-risk & downtime risk & production delay & allocation
shortfall \\
Criticality & asset priority & product priority & end-use criticality \\
\bottomrule
\end{tabular}
\end{table*}

Directionality is normalized so larger values indicate greater risk.
Public datasets provide statistical calibration. Causal graphs and
counterfactual labels come exclusively from the controlled benchmark
(\S V).

\subsection{Problem Formulation}

For archetype $a$, represent the supply chain as a temporal directed
acyclic graph (DAG) $G^{(a)}=(V^{(a)},E^{(a)},\Lambda^{(a)})$, where
$\Lambda^{(a)}$ stores edge-lag assignments. Detector $D_\psi$ emits per-timestep scores.
The first threshold crossing is the detection time. It is calibrated to a
false-alarm rate (FAR) $\leq$ 1/month.

For anomaly $e=(k,t,s_{k,t})$, where $k$ is the flagged KPI, $t$ its
detection time, and $s_{k,t}$ its score, candidates are upstream ancestors
of $k$ in the candidate graph $G_c$. In v1 the detection target is not an
arbitrary KPI: an episode enters the ranking study only when the flagged KPI is
the archetype's service-risk node (Table~\ref{tab:kpi}), so $k$ always denotes
that node. Candidates are then

\begin{equation}
C(e)=\{i\in V:i\rightsquigarrow_{G_c} k\}.
\end{equation}
Here $G_c$ is benchmark $G$ in the headline regime, recovered $\hat G$ in
the graph-swap regime, and an explicit all-non-$k$ construction without graph
features in the no-graph regime. Stored reported runs have nonempty benchmark
and recovered ancestor sets. For completeness, an empty set falls back to all
non-$k$ nodes. $C(e)$ is therefore the service-risk node's ancestor set: 11/9/8
candidates under benchmark $G$ for DI/SE/CM. Temporal precedence and lag
residual are features, not filters.

Each candidate receives an ordinal score from a learned ranker $R_\theta$.
The ranker is trained to order candidates using \textbf{net intervention
value} as relevance:

\begin{equation}
\widetilde{Y}_{i,e}=B_{i,e}\phi_{i,e}-c_{i,e},\qquad
Y_{i,e}=\max\{0,\widetilde{Y}_{i,e}\}.
\end{equation}

where:
\begin{itemize}
\item $B_{i,e}=L_e^{(0)}-L_e^{(i)}$ is paired loss avoided;
\item $\phi_{i,e}=\operatorname{clip}(1-\gamma
|\Delta_{i,e}|/\sigma_i,\phi_{\min},1)$, where
$\Delta_{i,e}=x_i(t)-\mu_i$, $\sigma_i$ is the v1 calibration-scale
standard deviation, and the main setting is $\gamma=0.75$ and
$\phi_{\min}=0.1$;
\item $c_{i,e}=c_aL_e^{(0)}/d_G(i,k)$, where $c_a$ is the archetype cost
scale, $L_e^{(0)}$ is no-action loss, and $d_G$ is directed shortest-path
length in benchmark $G$; unreachable nodes use the implemented surrogate
$d_G=|V|$.
\end{itemize}

$\widetilde{Y}_{i,e}$ is forced-action net value. The ranking label $Y_{i,e}$
truncates negative raw values to create a zero-valued no-action reference;
clipping does not implement test-time abstention. Score-based abstention is
evaluated separately in \S VI-F. The LambdaMART score orders candidates; it
is not a calibrated estimate of $Y_{i,e}$. In the no-graph regime, causal
strength and lag use fixed uninformative defaults 0.1 and 1 step for every
candidate.

The inverse-distance term is a stylized proxy for intervention type
rather than coordination distance. Candidates closer to the service-risk
node typically represent downstream buffering, emergency capacity, or
allocation actions requiring sustained resource commitment, whereas
upstream process corrections are modeled as lower-cost one-time actions.
Accordingly, the benchmark assigns higher cost to candidates closer to
the service-risk node. We do not interpret graph distance as an empirical
monetary-cost model. The pre-specified scale is $c_a=0.3$ for DI/SE and
$c_a=0.1$ for CM; sensitivity is
reported in \S VI.

From detection time onward, \textbf{do(stabilize \emph{i})} hard-clamps
KPI $i$ to baseline mean $\mu_i$. This is the primary benchmark label
specification. Factual and counterfactual rollouts
share a seed, so $Y_{i,e}$ is paired. Because this idealization is strong,
we treat partial, delayed, and stochastic responses as assumption stress
tests.

\section{DACRI Framework}

\subsection{Anomaly Detection}

We compare exponentially weighted moving average plus median absolute
deviation (EWMA+MAD), Isolation Forest, a long short-term memory
autoencoder (LSTM-AE), UnSupervised Anomaly Detection (USAD), and
multivariate time-series anomaly detection with graph attention networks
(MTAD-GAT). Validation thresholds target FAR $\leq$ 1/month and prioritize
lead time.

\subsection{Lagged Causal-Structure Estimation}

We compare Granger, PCMCI/PCMCI+ (primary), and a nonlinear check
\cite{b14,b15,b16}. Intrinsic recovery and downstream policy utility are
reported separately. The reported analysis configuration pre-specifies
$\alpha_{\rm PCMCI}=0.01$ before validation and test evaluation.

\subsection{Decision-Aware Intervention Ranking}

The ranker scores each candidate using: anomaly severity,
causal-strength statistic and lag, temporal precedence, lag-alignment
residual, downstream criticality, lifecycle-recovery feasibility,
and inventory-buffer depletion. Relevance labels come from paired
simulations. The headline ranking experiment uses benchmark $G$ to
isolate ranking; a separate graph-swap retrains under recovered $\hat G$
and no graph.

\textbf{LambdaMART} (LightGBM) is the episode-adaptive learned policy. We
compare it against a conceptual hierarchy. \textbf{Severity-only} is an
intentionally information-light monitoring baseline. \textbf{True-root}
is an attribution-as-action baseline that acts on the injected source.
\textbf{Train-selected
static} chooses the single training-set node with highest average net
value and reuses it for every test episode. Because the selection uses
simulator-derived counterfactual labels for every training candidate, it is
a strong full-information non-adaptive benchmark rather than an ordinary
logged-data operational policy. \textbf{Constant-buffer} always intervenes on the
archetype-specific loss-driving buffer. It is a strong, domain-informed,
oracle-like structural reference because that node is supplied in
advance. A \textbf{perfect-value oracle} selects the highest-value option in
$C(e)\cup\{\text{no action}\}$, with no-action value zero, and provides an
upper bound using realized counterfactual value. Equal-weight is
retained as an adaptive heuristic outside the six-level hierarchy.

\section{CriticalSCM-Bench v1}\label{sec:bench}

CriticalSCM-Bench combines a hand-designed temporal DAG, autoregressive
trajectories with seasonality/noise/missingness, sampled disruptions, and
paired counterfactual labels. Calibration magnitudes draw from Backblaze drive
data, electronics lifecycle data, SECOM, and USGS; graph and intervention
ground truth remain synthetic.
Calibration fixes the marginals of input nodes only. Downstream
service-risk dynamics follow from graph propagation and are not
empirically calibrated: realized dispersion at those sinks exceeds the
calibrated standard deviation by roughly $590\times$ (DI), $302\times$
(SE), and $6\times$ (CM). Benchmark magnitudes at the loss-bearing nodes
should therefore be read as graph-determined, not data-derived.
Train/test base seeds are disjoint (200 episodes each). V1 collapses
sub-tier networks, omits multi-product/shared-resource competition, and
keeps graphs fixed within episodes.

\subsection{Benchmark Archetype Specification}

\textbf{Digital Infrastructure (DI).} 12 nodes, two-stage topology.
Drive reliability is calibrated from Backblaze \cite{b8}; lifecycle and
bill-of-material context comes from electronics data \cite{b9}. Disruptions: recycler yield drop
(15--40\%), fault-rate spike (1.5--3$\times$), lead-time extension (+5--21d).
Lags: 7--14d (yield$\rightarrow$buffer), 1--3d (buffer$\rightarrow$downtime).
For the 30-day horizon, $L_e=\sum_\tau w_q(\tau)d^+(\tau)$, where
$d^+=\max\{\text{downtime\_risk}-\mu_d,0\}$ and
$w_q(\tau)=\max\{q(\tau)-\mu_q+1,1\}$ weights asset priority $q$.

\textbf{Semiconductors (SE).} 10 nodes spanning fab and downstream.
UCI SECOM \cite{b10} calibration. Disruptions: quality-hold spike (2--5$\times$),
rework-yield drop (10--30\%), lead-time extension (+7--28d). Lags: 3--7d
(quality$\rightarrow$inventory), 2--5d (inventory$\rightarrow$delay). Over
21 days, $L_e=\lambda_s\sum_\tau w_p(\tau)S^+(\tau)+
\lambda_d\sum_\tau w_p(\tau)D^+(\tau)$, with $\lambda_s=\lambda_d=1$;
$S^+(\tau)$ and $D^+(\tau)$ are inventory below and delay above baseline;
$w_p(\tau)=\max\{p(\tau)-\mu_p+1,1\}$ weights product priority $p$.

\textbf{Critical Materials (CM).} 9 nodes. USGS \cite{b11} calibration.
Multi-week propagation (4--8 week purity-to-availability lag).
Disruptions: purity decline (3--10 percentage points), export restriction (20--50\%),
stockpile draw (30--60\%). Over 26 weeks,
$L_e=\sum_\tau w_c(\tau)S^+(\tau)$, where $S^+(\tau)$ is allocation
shortfall above baseline and $w_c(\tau)=\max\{c(\tau)-\mu_c+1,1\}$ weights
end-use criticality $c$.

\textbf{Cross-Archetype Transfer.} Feature alignment uses quantile matching
within seven aligned KPI classes (inverse-CDF mapping); lag rescaling by median injected
lag; loss rescaling to zero mean/unit variance; frozen ranker
evaluation. Reported: DI$\rightarrow$SE and DI$\rightarrow$CM.

\section{Experiments}

The main ranking study uses 5 seeds with 200 disjoint train and test
episodes per archetype. Unless stated otherwise, seed-level point estimates
are medians. Reported seed-level intervals are interquartile ranges (IQRs),
whereas paired episode-level analyses use bootstrap confidence intervals
(CIs). Secondary-study budgets are stated
below. Because the ranking question is separate from detector selection,
the main study holds detection timing at the controlled proxy
$t_{\rm det}=t_{\rm inj}+5$. This is not a detector-specific threshold
crossing; the timing-sensitivity study sweeps the offset. Ranking quality
uses normalized discounted cumulative gain at 5 (NDCG@5), while the
separate detector study uses a FAR budget of 1 event/month.

\subsection{Decision-Aware Ranking}

Table~\ref{tab:ranking} is the principal decision-ranking comparison.

\begin{table*}[!tbp]
\caption{Decision-aware ranking (5 seeds $\times$ 200 test episodes). Cells:
median normalized net value / NDCG@5; bold marks the best non-oracle net
value. Values are normalized by no-action loss and rounded to three decimals;
text percentages use unrounded summaries. Labels floor negatives at zero.
\S\ref{subsec:feasibility} reports
forced-action and score-abstention accounting.}
\label{tab:ranking}
\centering\footnotesize\setlength{\tabcolsep}{3.5pt}
\begin{tabular}{@{}lccc@{}}
\toprule
Policy & DI & SE & CM \\
\midrule
LambdaMART & 0.075 / 0.267 & \textbf{0.049} / 0.249 &
\textbf{0.050} / 0.569 \\
Constant-buffer & \textbf{0.079} / 0.244 & 0.047 / 0.217 &
0.045 / 0.393 \\
Train-selected static & 0.064 / 0.262 & 0.047 / 0.217 &
0.045 / 0.393 \\
Equal-weight & 0.077 / 0.274 & 0.048 / 0.246 &
0.039 / 0.155 \\
Severity-only & 0.005 / 0.100 & 0.003 / 0.103 &
0.016 / 0.420 \\
True-root & 0.000 / 0.083 & 0.000 / 0.098 &
0.024 / 0.550 \\
Perfect-value oracle & 0.131 / -- & 0.090 / -- & 0.057 / -- \\
\bottomrule
\end{tabular}
\end{table*}

The zero floor in Table~\ref{tab:ranking} acts as a perfect episode-level
no-action reference whenever a selected action has negative raw value. The
table therefore measures ranking relevance against that reference, not
realized value from a deployable act-or-abstain policy. Absolute normalized
values are small because the feasibility scalar floors a large share of
candidate-episodes; \S\ref{subsec:feasibility} quantifies this, bounds the
interpretation of magnitude, and gives the forced-action and score-abstention
accounting.

Relative to the full-information train-selected static benchmark, LambdaMART's
median gain is 16.2\%/5.7\%/11.1\% (DI/SE/CM). It does not dominate stronger
references: constant-buffer is 5.1\% better on DI and the equal-weight adaptive
heuristic 3.5\% better, both relative to LambdaMART, which leads only on SE and
CM and by modest margins. Table~\ref{tab:ranking} therefore supports a bounded
conclusion: episode adaptation adds value in some regimes, but greater model
complexity is not uniformly beneficial.

Paired bootstrap confidence intervals and sign-flip tests use 1,000 test
episodes with disjoint episode seeds across the five runs. They support LambdaMART over static on
SE (mean +.0015, 95\% CI
[.0003,.0027], $p=.015$) and CM (+.0062, [.0013,.0104], $p=.008$),
but not DI (+.0061, [-.0063,.0182], $p=.323$). Against
constant-buffer, DI is significantly worse (-.0055,
[-.0086,-.0028], $p<.001$), while paired tests favor LambdaMART on SE and CM.
Thus the learned model has statistically supported incremental benchmark
value on SE and CM, but not on DI; whether those gains justify operational
complexity is outside this study.

The most consistent result is the distinction between attribution and action.
Severity-only obtains
7.2\%/6.1\%/32.9\% of LambdaMART's value, while true-root is near zero on
DI/SE and below LambdaMART on CM. The true-root result is partly a
property of the feasibility model rather than of root-cause action as
such: a just-disrupted node is frequently driven to the feasibility floor,
contributing to near-zero DI value (\S\ref{subsec:feasibility}). LambdaMART reaches 57\%/55\%/88\% of
the perfect-value oracle. The buffer is value-optimal in only
19.5\%/18.0\%/9.5\% of episodes, so a fixed buffer is strong on average
without being episode-optimal.

\subsection{Detection and Causal-Structure Results}

Table~\ref{tab:detector} reports detector coverage and delay under the FAR constraint.

\begin{table*}[!tbp]
\caption{Detection at FAR $\leq$ 1/month (5 seeds $\times$ 200). Cells:
event rate / conditional / censored delay. Delays are within-seed medians
over detected/all events (misses = 120 steps), then medians across seeds;
units are days (DI/SE) or weeks (CM). Bold marks the frontier.}
\label{tab:detector}
\centering\footnotesize\setlength{\tabcolsep}{1.5pt}
\begin{tabular}{@{}
  >{\raggedright\arraybackslash}p{(\textwidth - 6\tabcolsep) * \real{0.1200}}
  >{\raggedright\arraybackslash}p{(\textwidth - 6\tabcolsep) * \real{0.2933}}
  >{\raggedright\arraybackslash}p{(\textwidth - 6\tabcolsep) * \real{0.2933}}
  >{\raggedright\arraybackslash}p{(\textwidth - 6\tabcolsep) * \real{0.2934}}@{}}
\toprule
\begin{minipage}[b]{\linewidth}\raggedright
Detector
\end{minipage} & \begin{minipage}[b]{\linewidth}\raggedright
DI rate / cond. / cens. (d)
\end{minipage} & \begin{minipage}[b]{\linewidth}\raggedright
SE rate / cond. / cens. (d)
\end{minipage} & \begin{minipage}[b]{\linewidth}\raggedright
CM rate / cond. / cens. (w)
\end{minipage} \\
\midrule

EWMA+MAD & 44\% / 6.0 / 120 & 48\% / 5.0 / 120 & 43\% / 5.0 / 120 \\
Iso. Forest & 28\% / 15.5 / 120 & 30\% / 11.0 / 120 & 62\% / 8.0 /
12.0 \\
\textbf{LSTM-AE} & \textbf{60\%} / \textbf{4.0} / \textbf{11.5} &
\textbf{88\%} / \textbf{2.0} / \textbf{2} & \textbf{69\%} / \textbf{3.0}
/ \textbf{8} \\
USAD & 48\% / 5.0 / 120 & 78\% / 3.0 / 4 & 55\% / 4.0 / 24.5 \\
\textbf{MTAD-GAT} & \textbf{57\%} / \textbf{4.0} / \textbf{10} & 73\% /
2.0 / 3 & 55\% / 4.0 / 34.5 \\
\bottomrule
\end{tabular}
\end{table*}

On DI, LSTM-AE provides greater event coverage and substantially shorter
censored delay than EWMA under the same FAR budget. Detector selection should
therefore consider coverage and lead time jointly. The detector study uses a
longer monitoring window than the intervention-loss windows in \S V. CM EWMA detects fewer than
half the events in every seed, so its within-seed censored medians equal the
120-week horizon.

Using a PCMCI significance threshold $\alpha_{\rm PCMCI}=0.01$, PCMCI
recovers 0.75 / 0.44 / 0.67 of true edges on DI / SE
/ CM (medians over five seeds). Granger recovers 0.58 / 0.56 / 0.78 on the
same seeds, so Granger has higher recall on SE and CM and lower recall on
DI. In a controlled graph-swap (3 seeds $\times$ 150; ranker retrained
per regime), recovered $\hat G$ versus benchmark $G$ gives net value
0.060/0.060 (DI), 0.059/0.048 (SE), and 0.038/0.038 (CM); no graph gives
0.036/0.035/0.024. These secondary values are not directly comparable
to Table~\ref{tab:ranking}. Structural recall and downstream decision value are
different quantities, and the recovered graph does not consistently
reduce value here.

\subsection{Learning Curve}

With 25/50/100/200 training episodes (3 seeds, 200 test), LambdaMART
median net value is .028/.058/.067/.068 (DI),
.039/.041/.046/.052 (SE), and .016/.022/.023/.053 (CM). Medians improve
with sample count, but the path is not smooth. Seed-level trajectories
remain noisy, and CM changes sharply only at 200. The curves show that
the policy is sample-sensitive; they do not show convergence.

\subsection{Feature Ablation}

Leave-one-out (3 seeds $\times$ 150 episodes):

Lifecycle feasibility carries the clearest signal. Removing
\texttt{lifecycle\_feasibility} causes the largest median drop
(-.050/-.012/-.020 for DI/SE/CM); removing \texttt{edge\_lag} gives
-.028/-.015/-.013. Other removals are smaller or can improve value,
which suggests redundancy rather than a feature set in which every
variable is uniformly helpful.

\subsection{Cost-Scale Sensitivity}

We sweep $c_a\in\{.1,.2,.3,.5\}$ (3 seeds $\times$ 150).
LambdaMART exceeds constant-buffer only at .1 on DI, is equal or slightly
higher on SE, and leads CM at .1--.2 but loses at .3--.5. Buffer-optimal
frequency falls with cost. The cost model is an explicit benchmark factor,
not an empirical monetary estimate; the sweep evaluates policy dependence on
this factor. Monetary calibration remains external to v1.

\subsection{Feasibility and Intervention Sensitivity}\label{subsec:feasibility}

The feasibility scalar is a synthetic intervention-effect proxy, not an
empirical recovery probability. We therefore rerun the injected-root
versus loss-optimal agreement audit under four paired specifications:
no attenuation ($\phi=1$), mild attenuation ($\gamma=0.375$,
$\phi_{\min}=0.1$), the current setting ($\gamma=0.75$,
$\phi_{\min}=0.1$), and a higher floor ($\gamma=0.75$,
$\phi_{\min}=0.3$). Root/optimal disagreement spans
89.7--99.7\% (DI), 97.3--100\% (SE), and 57.0--74.3\% (CM), remaining
high across all four specifications with greater variation on CM.

The four specifications are not fully independent because floor saturation
makes portions of the attenuation sweep inert. Since $\phi$ uses calibrated
rather than larger realized dispersion (\S\ref{sec:bench}),
51.2\%/47.8\%/51.9\% of DI/SE/CM candidate-episodes are floored, rising to
87.8\% (DI) and 75.0\% (CM) for disrupted candidates. This contributes to
near-zero DI true-root value and limits interpretation of magnitude. The
audit supports persistence of root/action disagreement, not empirical
calibration of its size.

Table~\ref{tab:intervention} summarizes the effects of partial and delayed interventions.

\begin{table*}[!tbp]
\caption{Intervention sensitivity (3 seeds $\times$ 150 train/test). Cells:
LambdaMART / constant-buffer / severity-only normalized value; parentheses
show LambdaMART retention versus full clamp.}
\label{tab:intervention}
\centering\footnotesize\setlength{\tabcolsep}{2.5pt}
\begin{tabular}{@{}lccc@{}}
\toprule
Intervention & DI & SE & CM \\
\midrule
Full at detection & .060/.071/.002 (100\%) & .048/.047/.004 (100\%) &
.038/.026/.015 (100\%) \\
75\% effect & .045/.052/.002 (75\%) & .033/.032/.003 (68\%) &
.026/.023/.011 (70\%) \\
50\% effect & .020/.030/.001 (33\%) & .019/.017/.001 (39\%) &
.020/.017/.006 (52\%) \\
Five-step delay & .039/.056/.000 (64\%) & .030/.031/.003 (62\%) &
.025/.024/.006 (66\%) \\
75\% + delay & .030/.040/.000 (49\%) & .019/.020/.002 (40\%) &
.020/.020/.004 (54\%) \\
\bottomrule
\end{tabular}
\end{table*}

Partial effects alter value and ordering: constant-buffer remains stronger
on DI and becomes competitive on delayed SE. Policy ordering is therefore
not invariant to intervention fidelity. A further stochastic-response stress
test introduces efficacy variation, outright failure, delay, ramping, and
decay (5 regimes, 5 seeds). LambdaMART retention relative to the full-clamp
benchmark falls to 0.51/0.29/0.24 under the moderate regime and
0.13/0.00/0.06 under the weak temporary regime for DI/SE/CM.
No candidate has positive net value in 45--99\% of episodes, and CM's optimal node
changes in up to 63\%. Scenario-matched retraining on noisy labels does not
consistently outperform the frozen full-clamp ranker.

Clipped labels supply a zero-valued benchmark reference but assume exact
episode-level knowledge of whether an action's realized net value is negative.
They do not supply a deployable no-action rule. Forced
action has mean value $-0.13/-0.12/+0.00$ under full clamp (DI/SE/CM), and
roughly half of those episodes lack a positive-net candidate; under stochastic
response the share reaches 99\%. Separately for each archetype, seed, and
response regime, a training-selected rule chooses the top-score cutoff that
maximizes summed normalized raw net on training episodes, then freezes it for
test. Because the cutoff is selected separately for each assumed response
regime, this is diagnostic rather than a regime-agnostic deployment rule.
On full-clamp episodes, this rule retains 30--52\% of the corresponding
clipped-label headline value. Across stochastic regimes, it abstains in
71--100\% of episodes and remains approximately break-even (minimum mean
test value $-0.004$).

\subsection{Circular-Evaluation Control}\label{subsec:circular}

To test dependence on counterfactual labels, a ranker trained once on hard
clamps is frozen and evaluated under unseen intervention specifications
against scenario-matched retraining (5 seeds $\times$ 100 train/test).

Across the four mismatched idealizations, frozen/scenario-matched median
value ratios range from 1.02 to 1.13 on DI and from 0.93 to 1.21 on SE;
the CM mean across specifications is 0.83.
On CM, the frozen ranker loses about half its value relative to matched
retraining under five-step delay (ratio 0.49). Dependence on counterfactual
labels is therefore mild on DI and SE but material on CM, and the circular-evaluation
concern is reduced rather than removed: label mismatch, OOD, and transfer
tests all remain inside the same autoregressive simulator family.

\subsection{Out-of-Distribution Generalization}

In a paired leave-one-disruption-template-out OOD test, the same fitted model retains
95\%/92\%/62\% in-domain net value on DI/SE/CM (0.069$\rightarrow$0.065,
0.052$\rightarrow$0.048, 0.047$\rightarrow$0.029). NDCG retention is
96\%/95\%/78\%. Constant-buffer OOD net value is
0.072/0.046/0.049, exceeding LambdaMART on DI and CM. Thus CM has a
substantial held-out-disruption weakness.

\subsection{Schema Transfer}

Frozen DI$\rightarrow$SE and DI$\rightarrow$CM rankers retain 75\%
(0.036 vs 0.048) and 65\% (0.025 vs 0.038) of in-domain value. Transfer
is a secondary stress test. The larger CM drop is consistent with
several schema and time-scale differences. Because those differences
move together, this experiment does not isolate lag scale as the cause.

\subsection{Controlled Lag-Scale Test}

To probe the lag explanation directly, we ran a controlled test that
holds topology, weights, calibration, cost, disruptions, and the feature
schema fixed and varies only the edge-lag scale (factor
$f\in\{0.5,0.75,1.5,2.0\}$ about the trained regime), evaluating a frozen
ranker against one retrained at each $f$ (5 seeds). Across the four factors,
no monotone degradation pattern is observed. CM degrades only on the
stretch side---frozen-to-retrained value falls to 0.79/0.60 at
$f=1.5/2.0$, the day-to-week direction---while lag compression does not
hurt and DI/SE show no consistent effect. Thus lag scale alone does not
explain the cross-archetype transfer drop in this controlled test.

\subsection{Real-Data Detection Sanity Check}

We use Backblaze Drive Stats Q1 2024 (25.2M drive-days, 978 failures).
Isolation Forest detects pre-failure anomalies in 39\% of failed drives
at FAR 0.057 per 1k drive-days, median 21 days before failure; EWMA
detects 4\%. The Backblaze result supports the plausibility of early
detection and the detection-to-decision motivation; detector ordering differs
from simulation and therefore remains deployment-specific.

\subsection{Detection-Timing Sensitivity}

At detection offsets 3/5/10/14 time steps---days for DI/SE and weeks for
CM---(3 seeds $\times$ 150), LambdaMART value
is .059/.060/.011/.008 (DI), .059/.048/.027/.010 (SE), and
.079/.038/.013/.001 (CM). Because intervention begins at detection,
late flags leave less recoverable loss. Here the mechanism is direct:
detection timing changes the value that remains available to the policy.
The effect is severe, especially on CM.

\subsection{Illustrative Intervention Walkthrough}

Consider DI episode 2000001. A fault-rate spike is injected at drive
failure rate (\texttt{drive\_failure\_rate}) at $t=49$ and detected at
$t=54$; no-action
loss is 540.1. Severity-only selects recovery throughput
(\texttt{rework\_throughput}) and
true-root selects \texttt{drive\_failure\_rate}; both have zero net value.
LambdaMART selects spare-buffer level (\texttt{spare\_buffer\_level}),
obtaining 24.9 (0.046 normalized), while spare availability
(\texttt{spare\_availability}) obtains 16.2.
This episode is illustrative rather than general evidence, but it shows that
the largest anomaly, injected source, and best intervention can differ.

\subsection{Guarded Operator-Explanation Layer: Secondary Study}

This secondary study asks whether a language model can verbalize a fixed
DACRI decision, whether evidence constraints improve first-pass acceptance,
and whether validation provides fallback. Packets omit counterfactual labels,
and the schema has no recommendation field. The validator checks decision
preservation, packet entities, edges, lags and numbers, and rejects verified
root-cause or promised-outcome claims; rejected drafts use a packet template.

We build 90 packets (30 per archetype, 11 carrying a no-action
decision) and generate three times per packet under a generic and an
evidence-constrained prompt: 540 calls using the recorded API identifier
\texttt{deepseek-v4-flash}. Supplementary Artifact A contains generation
settings, prompt versions, packet schema, validator rules, and complete outputs.
For this explanation study only, no-action is locked when the top ranker
score falls below the pre-specified 10th percentile of an out-of-fold
training top-score distribution. This rule uses no test labels and is distinct
from the scenario-matched diagnostic cutoff in \S\ref{subsec:feasibility}.

\begin{table}[!b]
\caption{Guarded explanation layer (90 packets $\times$ 3 generations per
prompt; 540 calls), in percent. Draft acceptance precedes fallback; Guarded
includes 9.6\% template fallback. Grounding covers entities, edges, lags, and
numbers; stability requires identical text across generations.}
\label{tab:llm}
\centering\footnotesize\setlength{\tabcolsep}{2pt}
\begin{tabular}{@{}lcccc@{}}
\toprule
 & Template & Generic & Ev.-constr. & Guarded \\
\midrule
Decision preserved & 100 & 100 & 100 & 100 \\
Draft accepted & --- & 74.1 & 90.4 & 90.4 \\
Structurally grounded & 100 & 99.3 & 100 & 100 \\
Banned claim absent & 100 & 74.4 & 90.4 & 100 \\
Template fallback & 0 & --- & --- & 9.6 \\
Exactly stable over 3 gens & 100 & 0 & 0 & 2.2 \\
\bottomrule
\end{tabular}
\end{table}

No output redirected the fixed decision, and none of the 66 no-action outputs
recommended intervention.
Evidence-constrained prompting improves first-pass acceptance from 74.1\%
to 90.4\%, with 59 discordant
pairs favoring it against 15 favoring generic (exact McNemar
$p=2.55\times10^{-7}$). Evidence-constrained drafts achieve 100\%
structured grounding under the specified checks, compared with 99.3\%
for generic prompting; guarded outputs reach 100\% after deterministic
validation and fallback. All 26 rejections in the evidence-constrained
condition arise from one banned claim: describing the intervention target
as the verified root cause, an attribution--intervention conflation.

Generation stability is the weakest outcome: neither prompt yields three
identical generations for any packet; fallback raises exact stability only
to 2.2\%. Prose varies across generations, so deployment-oriented reporting
should disclose this variability. Checks cover structured claims and numeric
tokens, not usefulness, style, or complete semantic faithfulness. All packets
are simulator-derived, and no human audit was completed; this remains a
constrained secondary component.

\section{Discussion}

Attribution is not equivalent to intervention selection: severity-only and
true-root miss value, although feasibility-floor saturation partly drives the
true-root result. Adaptation is statistically supported over the
full-information static benchmark on SE and CM, but not DI, where
constant-buffer wins. Operational complexity was not evaluated. The oracle
gap leaves further policy headroom.

The stress tests map the benchmark's validity envelope. Intervention fidelity,
cost, timing, and graph specification alter value and ordering, while CM is
the weakest held-out-disruption regime. These are explicit, replaceable
benchmark factors rather than hidden deployment estimates: the results locate
where adaptive ranking helps within the controlled environment and where
structural policies remain preferable.

CriticalSCM-Bench supports controlled comparative evaluation, not deployment
certification. Independent simulators and retrospective operational data must
test transportability, monetary costs, and organizational feasibility. The
guarded explanation study evaluates a constrained renderer's safety and grounding, not
human factors.

\textbf{Artifact availability.} Supplementary Artifact A---the benchmark, the
evaluation harness, configurations, seeds, evaluation scripts, prompts,
validator rules, and every stored output behind the reported numbers---is
available at

{\centering\footnotesize\url{https://github.com/dyshang/dacri-criticalscm-bench}\par}

\section{Conclusion}

DACRI starts from a simple distinction: finding the source of a
disruption is not the same as choosing where to act. CriticalSCM-Bench
makes that distinction measurable through a paired, cost-aware
objective. Within the benchmark, severity and true-root often miss
recoverable value. LambdaMART improves a full-information static benchmark on SE and CM,
but loses to a strong structural rule on DI. Partial, delayed, cost, and OOD
tests identify regimes with statistically supported benchmark gains from
episode-specific ranking and those favoring structural rules. The contribution is a falsifiable
benchmark for deciding where to act, not universal learned-policy superiority.
Independently implemented simulators and retrospective operational data provide
the next tests of transportability.

\end{document}